\pdfoutput=1

\documentclass[11pt]{article}

\usepackage[svgnames]{xcolor}
\usepackage[final]{acl}
\newcommand{\highlight}[3][black]{{\fboxsep1.5pt\colorbox{#2}{\color{#1} #3}}}
\usepackage{times}
\usepackage{latexsym}
\usepackage{multirow}
\usepackage{adjustbox}
\usepackage{enumitem}
\usepackage[T1]{fontenc}

\usepackage[utf8]{inputenc}

\usepackage{microtype}

\usepackage{inconsolata}

\usepackage{graphicx}

\usepackage{amsmath}
\usepackage{amssymb}
\usepackage{todonotes}
\usepackage{xspace}

\newcommand{\amps}{\textsc{Amps}\xspace}
\newcommand{\ampstau}{\textsc{Amps$_\tau$}\xspace}
\newcommand{\Y}{\ensuremath{\mathbf{Y}}\xspace}
\newcommand{\X}{\ensuremath{\mathbf{X}}\xspace}

\title{\amps: ASR with  Multimodal Paraphrase Supervision}

\author{Abhishek Gupta$^*\quad$ Amruta Parulekar$^*\quad$ Sameep Chattopadhyay$\quad$ Preethi Jyothi  
\\ 
Indian Institute of Technology Bombay, Mumbai, India \\ 
\small{ \texttt{\{abhishekumgupta,amrutaparulekar.iitb,sameep.ch.2002\}@gmail.com, pjyothi@cse.iitb.ac.in}} 
} 

\begin{document}
\maketitle
\begin{abstract}
Spontaneous or conversational multilingual speech presents many challenges for state-of-the-art automatic speech recognition (ASR) systems. In this work, we present a new technique \amps that augments a multilingual multimodal ASR system with paraphrase-based supervision for improved conversational ASR in multiple languages, including Hindi, Marathi, Malayalam, Kannada, and Nyanja. We use paraphrases of the reference transcriptions as additional supervision while training the multimodal ASR model and selectively invoke this paraphrase objective for utterances with poor ASR performance. Using \amps with a state-of-the-art multimodal model SeamlessM4T, we obtain significant relative reductions in word error rates (WERs) of up to $5\%$. We present detailed analyses of our system using both objective and human evaluation metrics.
\end{abstract}

\section{Introduction}
 \renewcommand{\thefootnote}{*}
\footnotetext{These authors contributed equally to this work.}
\renewcommand{\thefootnote}{\arabic{footnote}}
Automatic speech recognition (ASR) systems have shown considerable progress in recent years but still falter when subjected to spontaneous conversational speech containing disfluencies, loosely articulated sounds, and other noise factors~\cite{gabler2023reconsidering}. This degradation in ASR performance could be largely attributed to the unavailability of labeled spontaneous speech in most languages. How can we effectively utilize the limited quantities of existing labeled spontaneous speech? Towards this, we propose \amps (\textbf{A}SR with \textbf{M}ultimodal \textbf{P}araphrase \textbf{S}upervision) that augments an existing multilingual multimodal ASR system with paraphrase-based supervision to improve ASR performance on spontaneous speech in multiple languages.

Unlike standalone ASR models that are exclusively trained to perform ASR, multimodal models (such as SpeechT5 \cite{ao-etal-2022-speecht5}, MAESTRO \cite{Chen2022MAESTROMS}, etc.) are trained on multiple tasks \emph{including ASR} using speech and text data in various paired (and unpaired) forms. We focus on one such multilingual multimodal model, SeamlessM4T ~\cite{seamlessm4tmassivelymultilingual}, that consists of dual encoders for speech and text and a shared text decoder, thus creating both speech-to-text and text-to-text pathways.%


\amps \footnote{Code for AMPS is available at \href{https://github.com/csalt-research/amps-asr}{https://github.com/csalt-research/amps-asr}.} leverages the multimodal nature of SeamlessM4T by introducing a paraphrasing objective jointly with ASR. Along with using spontaneous speech and its corresponding transcription to train the speech-to-text pathway in SeamlessM4T, \amps also uses paraphrases of the reference transcriptions as additional supervision to train the text-to-text pathway. We selectively employ paraphrase-based augmentation during training when the ASR loss is high (as determined by a predetermined threshold); high ASR loss is typically triggered by noise or poorly enunciated words in spontaneous speech. This selective intervention offers the model an alternate path of opting for semantically close words and phrases when the audio is not very clear. It is important that the paraphrases should not significantly differ in word order from the original transcripts, thus enabling the model to easily align representations of speech, text, and its paraphrase.

With \amps, we derive significant improvements in ASR for spontaneous speech in Hindi, Marathi, Malayalam, Kannada, and Nyanja compared to strong ASR-only finetuned baselines. We report improvements not only in terms of word error rate (WER) reductions but also using semantic evaluation metrics. We also conduct a detailed human evaluation comparing the outputs of \amps with the outputs from finetuning only with the ASR objective and show consistent improvements in human scores. We also present many ablations, including different paraphrasing techniques, the influence of varying thresholds on the performance of \amps, and using varying amounts of training data. We envision that techniques like \amps could be used to improve ASR of atypical speech for people with speech impairments where comprehensibility of the transcripts is critical (more than faithfulness of transcripts to the underlying speech, as highlighted in very recent work by~\citet{10447177}).

\vspace{-3pt}
\section{Related Work}
\vspace{-3pt}
In recent years, multimodal models for speech recognition have gained significant recognition \cite{ao-etal-2022-speecht5,Chen2022MAESTROMS,rubenstein2023audiopalmlargelanguagemodel,zhang2023speechgptempoweringlargelanguage}. These models are capable of processing both speech and text inputs and can be adapted for tasks such as translation and speech generation. A notable example is Meta AI's SeamlessM4T~\cite{seamlessm4tmassivelymultilingual}, which can support nearly 100 languages. One of the key advantages of such models is their ability to exploit text-only training to fine-tune shared parameters in the ASR pipeline. 
Some of the recent work on text-based adaptation for ASR models include \citet{vuong-etal-2023-adabert,Bataev2023TextonlyDA,10389682,mittal2023insitu}. One potential approach for leveraging text-only data for ASR finetuning is through training the text decoder with a paraphrasing objective. Emerging research \cite{Yu2023TrainingWT} has shown that text paraphrasing can be used to augment LLM performance but we are the first to show how paraphrases can be used to improve ASR. \citet{10447177} is a recent study focusing on meaning preservation in disordered speech transcription, but do not offer any technique to help improve meaning preservation in ASR outputs.
\section{Methodology}
\amps scaffolds on a multimodal base model comprising a speech encoder, a text encoder, and a shared decoder that takes inputs from both encoders. SeamlessM4T is an example of such a model, capable of performing multiple tasks including text-to-text translation (T2T), and speech-to-text transcription/translation (S2T).
\begin{figure}[t] 
  \includegraphics[width=\linewidth]{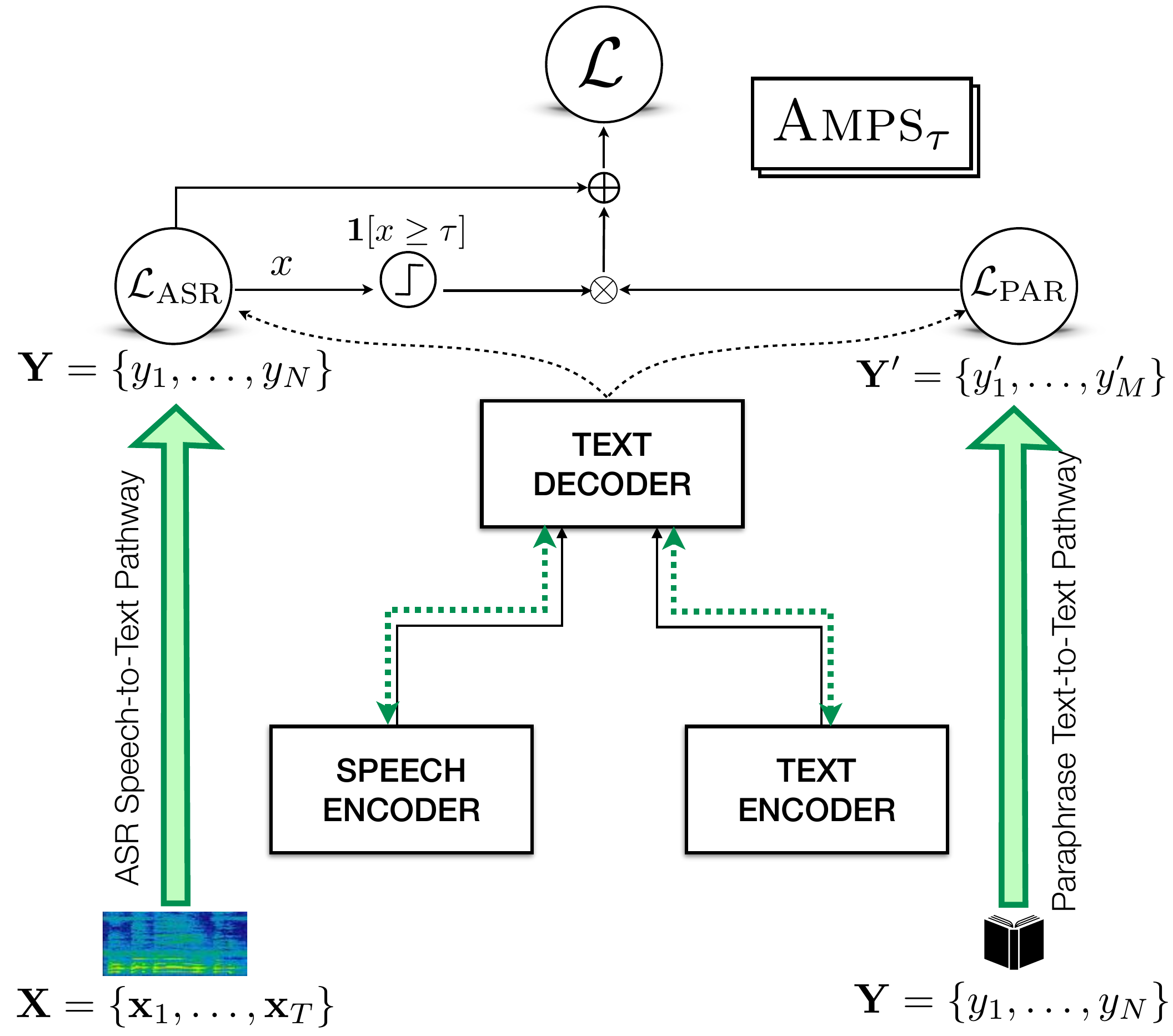}
  \caption{\textbf{Multimodal \ampstau Pipeline.} \ampstau applies a dual pass through the S2T pipeline with an ASR objective and the T2T pipeline with a paraphrasing objective. The paraphrasing loss is only incorporated when the ASR loss exceeds a predefined threshold. }
  \label{fig:amps}
  \end{figure}
We introduce a new auxiliary task of text-to-text paraphrasing. This allows the model to predict words that are semantically similar and fit within the context of the sentence, without significantly altering its word order. The shared decoder architecture of SeamlessM4T allows us to exploit common parameters of both S2T and T2T pipelines and enhance the ASR performance of the model.

Formally, consider a speech utterance \( \X = \{ \mathbf{x}_1, \mathbf{x}_2, \ldots, \mathbf{x}_L \,|\, \mathbf{x}_i \in \mathbb{R}^d \} \) with its corresponding transcript \( \mathbf{Y} = \{ y_1, y_2, \ldots, y_N \} \). For a transcript $\Y$, we generate a paraphrase $\Y'=\{ y'_1, y'_2, \ldots, y'_{M} \}$. Given a labeled instance $\{\X, \Y, \Y'\}$, the ASR, paraphrase, and the \amps loss functions are as follows.
\vspace{-1pt}
\begin{align}
\mathcal{L}_{\text{ASR}} &= \sum_{t=1}^{N} \log p_{\theta}(y_t \mid y_{<t}, \X), \nonumber \\
\mathcal{L}_{\text{PAR}}  &= \sum_{t=1}^{M} \log p_{\phi}(y'_t \mid y'_{<t}, \Y), \nonumber \\
\mathcal{L}_{\text{\amps}} &= \mathcal{L}_{\text{ASR}} + \mathcal{L}_{\text{PAR}}. \nonumber
\end{align}
\vspace{-1pt}
For each batch, we pass the audio through the S2T pathway and compute the ASR loss between the predicted and ground-truth transcriptions. We also pass the ground-truth transcriptions as input through the T2T pathway with paraphrase-based supervision to compute $\mathcal{L}_{\text{PAR}}$. Figure~\ref{fig:amps} illustrates a schematic of our proposed architecture. 

%
\begin{table*}[t!]
\centering
\begin{adjustbox}{max width=\textwidth}
\begin{small}
\renewcommand{\arraystretch}{0.65}
\begin{small}

\begin{tabular}{l|c|c|ccc|cc|c|c}
\hline
\multirow{4}{*}{Language} &
\multicolumn{1}{|c|}{\multirow{2}{*}{Evaluation Type}} &
  \multicolumn{1}{|c|}{\multirow{2}{*}{Direct Inference}} &
  \multicolumn{3}{c|}{\multirow{2}{*}{All Data}} &
  \multicolumn{2}{c|}{\multirow{2}{*}{Hard 100}} &
\multicolumn{2}{c}{\multirow{2}{*}{$\Delta = \ampstau -$ ASR}} 
   
\\
 
&
  \multicolumn{1}{|c|}{} &
  \multicolumn{1}{|c|}{} &
  \multicolumn{3}{c|}{} &
  \multicolumn{2}{c|}{} &
 
  \multicolumn{1}{c}{} &
  \multicolumn{1}{c}{} 
  \\ 

\cline{2-10}
 &
  \multicolumn{1}{|c|}{\multirow{2}{*}{Configuration}} &
  \multicolumn{1}{c|}{\multirow{2}{*}{-}} &
    \multicolumn{1}{c}{\multirow{2}{*}{ASR}} &
    \multicolumn{1}{c}{\multirow{2}{*}{\amps}} &
    \multicolumn{1}{c|}{\multirow{2}{*}{\ampstau}} &
 \multicolumn{1}{c}{\multirow{2}{*}{ASR}} &
  \multicolumn{1}{c|}{\multirow{2}{*}{\ampstau}} &
\multicolumn{1}{c}{\multirow{2}{*}{ $\Delta$Hard }} &
  \multicolumn{1}{c}{\multirow{2}{*}{ $\Delta$All }} 
\\ &
  \multicolumn{1}{|c|}{} &
\multicolumn{1}{|c|}{} &
\multicolumn{1}{c}{} &
\multicolumn{1}{c}{} &
\multicolumn{1}{c|}{} &
\multicolumn{1}{c}{} &

\multicolumn{1}{c|}{} &
\multicolumn{1}{c}{}  &
 \multicolumn{1}{c}{} 
 
   \\ \hline
\multirow{6}{*}{Marathi} &
  \multirow{2}{*}{WER $\downarrow$} &
  \multirow{2}{*}{38.65} &
  \multirow{2}{*}{21.18} &
  \multirow{2}{*}{21.58} &
  \multirow{2}{*}{\highlight{green!25}{\textbf{20.20}}} &
  \multicolumn{1}{c}{\multirow{2}{*}{48.91}} &
  \multicolumn{1}{c|}{\multirow{2}{*}{42.79}} &

  \multicolumn{1}{c|}{\multirow{2}{*}{-6.12}} &
\multicolumn{1}{c}{\multirow{2}{*}{-0.98}}
\\ 
 &
   &
   &
   &
   &
   &
   
   &
   &\\ 

  
 \multirow{1}{*}{}  &
  \multirow{2}{*}{METEOR $\uparrow$} &
    \multirow{2}{*}{59.84} &
  \multirow{2}{*}{73.32} &
  \multirow{2}{*}{\highlight{green!25}{\textbf{77.67}}} &
  \multirow{2}{*}{76.62} &
   \multicolumn{1}{c}{\multirow{2}{*}{54.13}} &
  \multicolumn{1}{c|}{\multirow{2}{*}{58.45}} &
 
  \multicolumn{1}{c|}{\multirow{2}{*}{4.32}} &
\multicolumn{1}{c}{\multirow{2}{*}{3.30}}
\\ 
 &
   &
   &
   &
   &
  
   &
   &
   &\\ 
  &
  \multirow{2}{*}{BERTScore $\uparrow$} &
    \multirow{2}{*}{81.01} &
  \multirow{2}{*}{90.40} &
  \multirow{2}{*}{\highlight{green!25}{\textbf{92.31}}} &
  \multirow{2}{*}{91.92} &
   \multicolumn{1}{c}{\multirow{2}{*}{84.73}} &
  \multicolumn{1}{c|}{\multirow{2}{*}{85.82}} &

  \multicolumn{1}{c|}{\multirow{2}{*}{0.99}} &
\multicolumn{1}{c}{\multirow{2}{*}{1.52}}
\\ 
 &
   &
   &
   &
   &
   
   &
   &
   &\\ \hline
  
  \multirow{6}{*}{Hindi} &
  \multirow{2}{*}{WER $\downarrow$} &
  \multirow{2}{*}{29.16} &
  \multirow{2}{*}{20.63} &
  \multirow{2}{*}{20.83} &
  \multirow{2}{*}{\highlight{green!25}{\textbf{20.12}}} &
 \multicolumn{1}{c}{\multirow{2}{*}{49.09}} &
  \multicolumn{1}{c|}{\multirow{2}{*}{45.91}} &

  \multicolumn{1}{c|}{\multirow{2}{*}{-3.18}} &
\multicolumn{1}{c}{\multirow{2}{*}{-0.51}}
\\ 
 &
   &
   &
   &
   &
   
   &
   &
   &\\ 
  
  
 \multirow{1}{*}{}  &
  \multirow{2}{*}{METEOR $\uparrow$ } &
    \multirow{2}{*}{72.25} &
  \multirow{2}{*}{81.04} &
  \multirow{2}{*}{81.38} &
  \multirow{2}{*}{\highlight{green!25}{\textbf{81.56}}} &
    \multicolumn{1}{c}{\multirow{2}{*}{57.66}} &
  \multicolumn{1}{c|}{\multirow{2}{*}{60.91}} &
 
  \multicolumn{1}{c|}{\multirow{2}{*}{3.25}} &
\multicolumn{1}{c}{\multirow{2}{*}{0.52}}
\\ 
 &
   &
   &
   &
   &
  
   &
   &
   &\\ 
  &
  \multirow{2}{*}{BERTScore $\uparrow$} &
    \multirow{2}{*}{88.55} &
  \multirow{2}{*}{93.60} &
  \multirow{2}{*}{93.65} &
  \multirow{2}{*}{\highlight{green!25}{\textbf{93.76}}} &
  \multicolumn{1}{c}{\multirow{2}{*}{84.46}} &
  \multicolumn{1}{c|}{\multirow{2}{*}{85.44}} &

  \multicolumn{1}{c|}{\multirow{2}{*}{0.98}} &
\multicolumn{1}{c}{\multirow{2}{*}{0.16}}
\\ 
 &
   &
   &
   &
   &
   
   &
   &
   &\\ \hline
  \multirow{6}{*}{Malayalam} &
  \multirow{2}{*}{WER $\downarrow$} &
  \multirow{2}{*}{56.15} &
  \multirow{2}{*}{42.06} &
  \multirow{2}{*}{42.09} &
  \multirow{2}{*}{\highlight{green!25}{\textbf{39.97}}} &
\multicolumn{1}{c}{\multirow{2}{*}{74.86}} &
  \multicolumn{1}{c|}{\multirow{2}{*}{64.66}} &
 
  \multicolumn{1}{c|}{\multirow{2}{*}{-10.2}} &
\multicolumn{1}{c}{\multirow{2}{*}{-2.09}}
\\ 
 &
   &
   &
   &
   &
  
   &
   &
   &\\ 
 
  
 \multirow{1}{*}{}  &
  \multirow{2}{*}{METEOR $\uparrow$} &
    \multirow{2}{*}{43.69} &
  \multirow{2}{*}{60.39} &
  \multirow{2}{*}{60.31} &
  \multirow{2}{*}{\highlight{green!25}{\textbf{62.01}}} &
 \multicolumn{1}{c}{\multirow{2}{*}{32.48}} &
  \multicolumn{1}{c|}{\multirow{2}{*}{40.58}} &
 
  \multicolumn{1}{c|}{\multirow{2}{*}{8.10}} &
\multicolumn{1}{c}{\multirow{2}{*}{1.62}}
\\ 
 &
   &
   &
   &
   &
  
   &
   &
   &\\ 
  &
  \multirow{2}{*}{BERTScore $\uparrow$} &
    \multirow{2}{*}{84.35} &
  \multirow{2}{*}{91.50} &
  \multirow{2}{*}{91.56} &
  \multirow{2}{*}{\highlight{green!25}{\textbf{92.02}}} &
 \multicolumn{1}{c}{\multirow{2}{*}{85.40}} &
  \multicolumn{1}{c|}{\multirow{2}{*}{87.41}} &
 
  \multicolumn{1}{c|}{\multirow{2}{*}{2.01}} &
\multicolumn{1}{c}{\multirow{2}{*}{0.52}}
\\ 
 &
   &
   &
   &
   &
  
   &
   &
   &\\ \hline
  \multirow{6}{*}{Kannada} &
  \multirow{2}{*}{WER $\downarrow$} &
  \multirow{2}{*}{69.29} &
  \multirow{2}{*}{41.41} &
  \multirow{2}{*}{40.10} &
  \multirow{2}{*}{\highlight{green!25}{\textbf{39.50}}} &
\multicolumn{1}{c}{\multirow{2}{*}{72.23}} &
  \multicolumn{1}{c|}{\multirow{2}{*}{67.58}} &
  
  \multicolumn{1}{c|}{\multirow{2}{*}{-4.65}} &
\multicolumn{1}{c}{\multirow{2}{*}{-1.91}}
\\ 
 &
   &
   &
   &
   &
  
   &
   &
   &\\ 
   
 
  
 \multirow{1}{*}{}  &
  \multirow{2}{*}{METEOR $\uparrow$} &
    \multirow{2}{*}{31.13} &
  \multirow{2}{*}{60.84} &
  \multirow{2}{*}{61.27} &
  \multirow{2}{*}{\highlight{green!25}{\textbf{61.68}}} &
  \multicolumn{1}{c}{\multirow{2}{*}{33.44}} &
  \multicolumn{1}{c|}{\multirow{2}{*}{38.30}} &
 
  \multicolumn{1}{c|}{\multirow{2}{*}{4.86}} &
\multicolumn{1}{c}{\multirow{2}{*}{0.84}}
\\ 
 &
   &
   &
   &
   &
  
   &
   &
   &\\ 
  &
  \multirow{2}{*}{BERTScore $\uparrow$} &
    \multirow{2}{*}{76.65} &
  \multirow{2}{*}{89.84} &
  \multirow{2}{*}{90.21} &
  \multirow{2}{*}{\highlight{green!25}{\textbf{90.41}}} &
   \multicolumn{1}{c}{\multirow{2}{*}{82.36}} &
  \multicolumn{1}{c|}{\multirow{2}{*}{85.54}} &
  
  \multicolumn{1}{c|}{\multirow{2}{*}{3.18}} &
\multicolumn{1}{c}{\multirow{2}{*}{0.57}}
\\ 
 &
   &
   &
   &
   &
   
   &
   &
   &\\ \hline
   
\end{tabular}%
 
\end{small}
\end{small}
    \end{adjustbox}
\caption{Comparing the performance of pure ASR, \amps, and \ampstau systems using 50 hours of training data with round-trip translated paraphrases. Best overall scores for each metric are highlighted in \highlight{green!25}{\phantom{xx}}.}
\label{tabmain}
\vskip -0.2 in
\end{table*}
\vspace{-1pt}
\paragraph{\ampstau: Loss Function Thresholding.} We aim at improving the model’s performance in noisy regions where the ASR loss is high by selectively triggering the paraphrase objective only when the ASR loss exceeds a predefined threshold $\tau$. \\
Thus, the loss for the system is given by
\begin{align}
\mathcal{L}_{\ampstau} &= 
\begin{cases} 
\mathcal{L}_{\text{ASR}} + \mathcal{L}_{\text{PAR}} & \text{if } \mathcal{L}_{\text{ASR}} > \tau, \\
\mathcal{L}_{\text{ASR}}  & \text{otherwise},
\end{cases}
\end{align}
where $\tau$ is a hyperparameter chosen based on ASR validation losses. Henceforth, \amps with the best threshold will be referred to as \ampstau. $\tau$ values for various experiments are in Appendix ~\ref{sec:thresh}. 

\section{Experimental Setup}
For all our experiments, we use the SeamlessM4T multilingual multimodal model~\cite{seamlessm4tmassivelymultilingual}. The text encoder and decoder modules are initialized using Meta's No Language Left Behind (NLLB) model \cite{nllbteam2022languageleftbehindscaling}. The speech encoder in SeamlessM4T uses Wav2Vec-BERT 2.0~\cite{Kessler2021Continualwav2vec2AA}, which is trained on over a million hours of unlabeled speech data. Further model details are in Appendix ~\ref{sec:modeldets}. 

\paragraph{Datasets.} The IndicVoices dataset~\cite{javed2024indicvoicesbuildinginclusivemultilingual} is a large collection of natural speech (74\% extempore, 17\% conversational and 9\% read) in 22 Indic languages. Among the languages we chose, Marathi, Kannada, and Malayalam are classified as low-resource by SeamlessM4T \cite{seamlessm4tmassivelymultilingual}, while Hindi is medium-resource. IndicVoices is the only multilingual open-source Indian speech corpus containing spontaneous speech and amongst the very few sources published after SeamlessM4T's release.%
\footnote{This dataset was chosen also to ensure that there was no data leakage between the SeamlessM4T training data and the evaluation sets.}
%
We also performed experiments on Nyanja (a low-resource language from Zambia) from the Zambezi-Voice dataset~\cite{sikasote23_interspeech}. 

We use roughly 50 hours of (predominantly conversational, henceforth referred to as \emph{mixed}) training data for each of the four Indian languages. For Hindi, we also simulate a very low-resource setting with random 5-hour samples of mixed and read training speech. For Nyanja, we used 5 hours of training data. (For Indic languages, our test sets are the validation sets that are part of IndicVoices. For Nyanja, we use the existing test set.) Given the limited amount of training data, we use parameter-efficient finetuning of adapter layers~\cite{PETNLP} in the speech encoder and text decoder layers of the SeamlessM4T model; more implementation details are in Appendix~\ref{sec:impl}.

\paragraph{Paraphrasing.} We translated the reference transcriptions into English using IndicTrans-2~\cite{gala2023indictrans} for the Indic languages and NLLB~\cite{nllbteam2022languageleftbehindscaling} for Nyanja before translating them back to their original languages.
For the Hindi mixed 5-hr setting, we experimented with top-$K$, $K=50$, and nucleus (top-$P$, $P=0.95$) sampling during round-trip translation to produce more diverse paraphrases. We also explored generating paraphrases using the multilingual LLM Aya-23~\cite{ayamodelinstructionfinetuned}. The exact prompt and other details are in Appendix~\ref{sec:prompt} and ~\ref{sec:para}. We used round-trip translation-based paraphrases for all the 50-hour experiments due to poor-quality LLM paraphrases for low-resource languages like Malayalam.

\paragraph{Evaluation Metrics.} Evaluation metrics used were Word Error Rate (WER), METEOR and the F1 score provided by BERTScore. More details are provided in Appendix \ref{sec:metrics}.
\begin{table*}[t]
\centering
\begin{adjustbox}{max width=\textwidth}
\begin{small}
 \renewcommand{\arraystretch}{1.2}
\begin{tabular}{l|c|c|ccc|c|cc|cc|cc}
\hline
\multirow{4}{*}{Language} &
\multicolumn{1}{|c|}{\multirow{1}{*}{Paraphrase}} &
  \multicolumn{1}{|c|}{\multirow{1}{*}{Direct}} &
  \multicolumn{3}{c|}{\multirow{1}{*}{Read Speech}} &
  \multicolumn{7}{c}{\multirow{1}{*}{Mixed Speech}}
\\ \cline{4-13}
 
  \multicolumn{1}{c}{}&
  \multicolumn{1}{|c|}{Type} &
  \multicolumn{1}{|c|}{Inference} &
  \multicolumn{3}{c|}{RT Trans} &
   \multicolumn{1}{c|}{} &
  \multicolumn{2}{c|}{RT Trans} &
  \multicolumn{2}{c|}{LLM-Para} &
   \multicolumn{2}{c}{TK+Nuc RT Trans}\\ 

\cline{2-13}
 &
  \multicolumn{1}{|l|}{\multirow{2}{*}{Configuration}} &
  \multicolumn{1}{c|}{\multirow{2}{*}{-}} &
    \multicolumn{1}{c}{\multirow{2}{*}{ASR}} &
    \multicolumn{1}{c}{\multirow{2}{*}{\amps}} &
    \multicolumn{1}{c|}{\multirow{2}{*}{\ampstau}} &
 \multicolumn{1}{c|}{\multirow{2}{*}{ASR}} &
    \multicolumn{1}{c}{\multirow{2}{*}{\amps}} &
    \multicolumn{1}{c|}{\multirow{2}{*}{\ampstau}} &

    \multicolumn{1}{c}{\multirow{2}{*}{\amps}} &
    \multicolumn{1}{c|}{\multirow{2}{*}{\ampstau}} &
  
    \multicolumn{1}{c}{\multirow{2}{*}{\amps}} &
    \multicolumn{1}{c}{\multirow{2}{*}{\ampstau}} 
 
\\ &
  \multicolumn{1}{|l|}{} &
\multicolumn{1}{|c|}{} &
\multicolumn{1}{c}{} &
\multicolumn{1}{c}{} &
\multicolumn{1}{c|}{} &
\multicolumn{1}{c|}{} &
\multicolumn{1}{c}{} &
\multicolumn{1}{c|}{} &

\multicolumn{1}{c}{} &
\multicolumn{1}{c|}{} &

\multicolumn{1}{c}{} &
\multicolumn{1}{c}{} 
 
   \\ \hline

\multirow{6}{*}{Hindi} &
  \multirow{2}{*}{WER $\downarrow$} &
  \multirow{2}{*}{29.16} &
  \multirow{2}{*}{\textbf{28.19}} &
  \multirow{2}{*}{28.94} &
  \multirow{2}{*}{28.57} &
  \multirow{2}{*}{23.14} &
  \multirow{2}{*}{23.14} &
  \multirow{2}{*}{\textbf{22.80}} &

  \multirow{2}{*}{22.35} &
   \multirow{2}{*}{\textbf{22.20}} &

  \multirow{2}{*}{\textbf{22.58}} &
  \multirow{2}{*}{22.81}
\\
 &
   &
   &
   &
   &
   &
   &
   &
   &
   &
   &
   & \\ \cline{2-13}
 
 
   
   &
  \multirow{2}{*}{METEOR $\uparrow$} &
  \multirow{2}{*}{72.25} &
  \multirow{2}{*}{\textbf{74.36}} &
  \multirow{2}{*}{73.58} &
  \multirow{2}{*}{73.91} &
  \multirow{2}{*}{79.10} &
  \multirow{2}{*}{78.86} &
  \multirow{2}{*}{\textbf{78.93}} &
 
  \multirow{2}{*}{79.25} &
   \multirow{2}{*}{\textbf{79.28}} &
 
  \multirow{2}{*}{\textbf{79.27}} &
  \multirow{2}{*}{79.11}
\\
 &
   &
   &
   &
   &
   &
  
   &
   &
   &
   & \\ \cline{2-13}
   &
  \multirow{2}{*}{BERTScore $\uparrow$} &
  \multirow{2}{*}{88.55} &
  \multirow{2}{*}{\textbf{90.39}} &
  \multirow{2}{*}{89.86} &
  \multirow{2}{*}{90.13} &
  \multirow{2}{*}{92.60} &
  \multirow{2}{*}{92.59} &
  \multirow{2}{*}{\textbf{92.78}} &

  \multirow{2}{*}{\textbf{92.89}} &
   \multirow{2}{*}{92.90} &

  \multirow{2}{*}{\textbf{92.63}} &
  \multirow{2}{*}{92.62}
\\
 &
   &
   &
   &
   &
   &
   
   &
   &
   &
   & \\ \hline   
\end{tabular}%
 
\end{small}
    \end{adjustbox}
\caption{Comparing ASR, \amps and \ampstau systems using 5 hours of mixed (conversational and read) speech with round-trip translations (RT Trans), LLM paraphrasing and top-K + nucleus paraphrasing.}
\label{tabsec}
\vskip -0.1 in
\end{table*}



\vspace{-3pt}
\section{Experiments and Results}
\vspace{-3pt}
Table~\ref{tabmain} shows the main results for all the 50-hour Indian-language experiments. $\ampstau$ consistently performs best compared to ASR, and the WER reductions are statistically significant (at $p < 0.05$ using the mapsswe test).%
\footnote{We also trained a variant where instances with a ASR loss were downweighted and instances with a high ASR loss were upweighted, thus forcing the model to focus more on the latter. This performed comparably to our baseline ASR model.}
%
Apart from the overall scores in \emph{All Data}, we sorted the transcriptions in descending order of WER using pure ASR and averaged metrics were calculated for both pure ASR and \ampstau for the first 100 (hardest) sentences. Improvements from ASR to \ampstau for these hardest 100 predictions are labeled \textit{$\Delta$Hard} in Table~\ref{tabmain}. We see that \textit{$\Delta$Hard} consistently exceeds\textit{ $\Delta$All}, indicating that the most improvement is observed in cases where pure ASR performs poorly. This supports the thresholding approach that triggers the paraphrase loss only when pure ASR predictions fall below a threshold. From our manual inspection of Hindi samples in the hardest-100 subset, we observe examples where pure ASR tends to produce acoustically similar but incorrect words, while \ampstau correctly identifies the words. For example, pure ASR misrecognized ``hua" (meaning 'is') as ``ugwa" (meaning 'grows') in a Hindi example; \ampstau gets this example right.

\begin{table}[t!]
\centering
\begin{adjustbox}{max width=0.5\textwidth}
\begin{small}
 
\begin{tabular}{l|ccc}
\hline
\multirow{2}{*}{Language} &
\multicolumn{1}{|c}{\multirow{2}{*}{ASR}} &
  \multicolumn{1}{c}{\multirow{2}{*}{\amps}} &
  \multicolumn{1}{c}{\multirow{2}{*}{\ampstau }}  
\\
 
  \multicolumn{1}{c}{}&
  \multicolumn{1}{|c}{} &
   \multicolumn{1}{c}{} &
  \multicolumn{1}{c}{} \\ \hline

\multirow{2}{*}{Marathi} &
  \multirow{2}{*}{4.199} &
  \multirow{2}{*}{4.271} &
  \multirow{2}{*}{\textbf{4.314}} 
\\ 
 
   &
   &
   &

   \\ \hline
\multirow{2}{*}{Hindi} &
  \multirow{2}{*}{3.608} &
  \multirow{2}{*}{3.625} &
  \multirow{2}{*}{\textbf{3.689}} 
\\ 
 
   &
   &
   &

   \\ \hline
   \multirow{2}{*}{Malayalam} &
  \multirow{2}{*}{3.635} &
  \multirow{2}{*}{3.688} &
  \multirow{2}{*}{\textbf{3.902}}  
\\ 
 
   &
   &
   &

   \\ \hline
   \multirow{2}{*}{Kannada} &
  \multirow{2}{*}{3.433} &
  \multirow{2}{*}{3.542} &
  \multirow{2}{*}{\textbf{3.597}} 
\\ 
 
   &
   &
   &

   \\ \hline

\end{tabular}%
 
\end{small}
    \end{adjustbox}
\caption{Comparison of human annotation results for ASR, \amps and \ampstau on a scale from 0 to 5.}
\label{tabhuman}
\end{table}
\subsection{Comparing Paraphrase Techniques}
Table~\ref{tabsec} shows results from training on 5 hrs of read/mixed Hindi speech and different paraphrasing techniques with mixed speech. Here, by mixed speech, we refer to a mixture of both read and conversational speech. Unsurprisingly, training on mixed speech yields significantly lower WERs compared to training on read speech. The highest performance gains were obtained using LLM paraphrasing for Hindi, suggesting that the LLM is a good option for medium-resource languages like Hindi. LLM outputs are subpar for low-resource languages like Kannada, and hence are not an option. Comprehensive results comparing the paraphrase techniques for other languages are given in Appendix \ref{sec:more1} and \ref{sec:more2}.

\subsection{Human Evaluation}
The transcription capabilities of ASR, \amps, and \ampstau models were verified through extensive human evaluation of the utterances with differing model outputs. The annotators reviewed 172, 153, 216, and 229 instances for Hindi, Marathi, Kannada, and Malayalam, respectively, giving a max score of 5 for a perfect transcript and penalizing them for minor errors (spellings, etc.) and major errors (incorrect semantics). The annotators were asked not to penalize a semantically identical word that differs from the speech. More details and scoring guidelines are provided in Appendix ~\ref{sec:human1} and qualitative examples are in Appendix~\ref{sec:human2}. Table~\ref{tabhuman} shows the averaged scores with \ampstau consistently performing the best across all languages.


\subsection{\amps for Nyanja}

Table~\ref{tabnonindic} shows overall results\footnote{Only WER and METEOR are reported. BERTScore does not support Nyanja.} on Nyanja with 5 hours of training data and round-trip translated paraphrases. Again, \ampstau performs the best, showing that \amps could be applied to diverse languages across language families.
%
\begin{table}[h!]
\centering
\begin{adjustbox}{max width=0.5\textwidth}
\begin{small}
 
\begin{tabular}{l|c|c|ccl}
\hline
\multirow{2}{*}{Language} &
\multicolumn{1}{|c|}{\multirow{2}{*}{Config.}} &
  \multicolumn{1}{c|}{\multirow{1}{*}{Direct}} &
  \multicolumn{1}{c}{\multirow{2}{*}{ASR}} &
  \multicolumn{1}{c}{\multirow{2}{*}{\amps}} &
  \multicolumn{1}{l}{\multirow{2}{*}{\ampstau}} 
\\
 
  \multicolumn{1}{c}{}&
  \multicolumn{1}{|c|}{} &
   \multicolumn{1}{|c|}{Inference} &
  \multicolumn{3}{c}{}   \\ \hline

\multirow{4}{*}{Nyanja} &
  \multirow{2}{*}{WER $\downarrow$} &
  
  \multirow{2}{*}{42.34} &
  \multirow{2}{*}{22.16} &
  \multirow{2}{*}{21.90} &
  \multirow{2}{*}{\textbf{21.59}} 
\\
 
   &
   &
   &
   &
   &
   
   \\ \cline{2-6}
  
 
  
\multirow{1}{*}{}&
  \multirow{2}{*}{METEOR $\uparrow$} &
  
  \multirow{2}{*}{66.71} &
  \multirow{2}{*}{79.25} &
  \multirow{2}{*}{79.30} &
  \multirow{2}{*}{\textbf{80.10}} 
\\
 
   &
   &
   &
   &
   &
  
   \\ \hline

\end{tabular}%
 
\end{small}
    \end{adjustbox}
\caption{Comparison of WER (\%) and METEOR for ASR, \amps and \ampstau for 5 hours Nyanja speech with round-trip translated paraphrases.}
\vspace{-1em}
\label{tabnonindic}
\end{table}

\subsection{Conclusion}
This work introduces a novel paraphrase-based supervision technique \amps to improve the ASR performance of spontaneous speech in multimodal models. This auxiliary supervision makes the model more robust and helps the model generalize better, especially in utterances with large ASR errors. We show significant ASR improvements on multiple and diverse languages and further validate these improvements via a thorough human evaluation. 

The broader idea of using textual supervision, as we did with paraphrases, to improve speech understanding is an interesting avenue to explore further. Future work will investigate how techniques like \amps could be used to improve ASR for atypical speech. Also, we used a predefined threshold on the ASR loss to trigger the paraphrase objective; this could be made a learnable quantity.

\section{Acknowledgements}

The authors thank the anonymous reviewers for their constructive feedback that improved the quality of the draft. The last author gratefully acknowledges support from the Amazon IITB AI ML Initiative.  

\vspace{-3pt}
\section*{Limitations}
\vspace{-3pt}
The primary limitation of our study was the lack of any appropriate pre-existing evaluation metric for the task. When supervising with paraphrases, the model often predicts semantically similar words or phrases that do not exactly match the transcript, making traditional metrics like Word Error Rate (WER) overly harsh for such cases. While BERTScore addresses semantic similarity, recent research suggests using LLMs to directly assess whether sentence meaning is preserved~\cite{10447177}.  In the future, we plan to adopt LLM-based evaluation alongside human reviews to improve assessment.

A second limitation was the occurrence of transliterated English words caused minor spelling errors in the model. We plan to mitigate this in the future by introducing code-switched words in our paraphrases to teach the model to associate the transliterated English words with their Latin script counterparts. Multilingual models like SeamlessM4T possess the unique ability to link semantically similar words across languages, thus comprehending code-switched speech easily and we aim to leverage this ability as future work.

Additionally, the threshold value $\tau$ is manually defined and not a dynamic value that is learned across languages. In future work, we plan to make this threshold a learnable parameter.

\bibliography{custom}

\begin{thebibliography}{30}
\providecommand{\natexlab}[1]{#1}

\bibitem[{Ao et~al.(2022)Ao, Wang, Zhou, Wang, Ren, Wu, Liu, Ko, Li, Zhang, Wei, Qian, Li, and Wei}]{ao-etal-2022-speecht5}
Junyi Ao, Rui Wang, Long Zhou, Chengyi Wang, Shuo Ren, Yu~Wu, Shujie Liu, Tom Ko, Qing Li, Yu~Zhang, Zhihua Wei, Yao Qian, Jinyu Li, and Furu Wei. 2022.
\newblock \href {https://doi.org/10.18653/v1/2022.acl-long.393} {{S}peech{T}5: Unified-modal encoder-decoder pre-training for spoken language processing}.
\newblock In \emph{Proceedings of the 60th Annual Meeting of the Association for Computational Linguistics (Volume 1: Long Papers)}, pages 5723--5738, Dublin, Ireland. Association for Computational Linguistics.

\bibitem[{Banerjee and Lavie(2005)}]{banerjee-lavie-2005-meteor}
Satanjeev Banerjee and Alon Lavie. 2005.
\newblock \href {https://aclanthology.org/W05-0909} {{METEOR}: An automatic metric for {MT} evaluation with improved correlation with human judgments}.
\newblock In \emph{Proceedings of the {ACL} Workshop on Intrinsic and Extrinsic Evaluation Measures for Machine Translation and/or Summarization}, pages 65--72, Ann Arbor, Michigan. Association for Computational Linguistics.

\bibitem[{Bataev et~al.(2023)Bataev, Korostik, Shabalin, Lavrukhin, and Ginsburg}]{Bataev2023TextonlyDA}
Vladimir Bataev, Roman Korostik, Evgeny Shabalin, Vitaly Lavrukhin, and Boris Ginsburg. 2023.
\newblock \href {https://api.semanticscholar.org/CorpusID:257219540} {Text-only domain adaptation for end-to-end asr using integrated text-to-mel-spectrogram generator}.
\newblock In \emph{Interspeech}.

\bibitem[{Chen et~al.(2023)Chen, Gong, and Qian}]{10389682}
Chang Chen, Xun Gong, and Yanmin Qian. 2023.
\newblock \href {https://doi.org/10.1109/ASRU57964.2023.10389682} {Efficient text-only domain adaptation for ctc-based asr}.
\newblock In \emph{2023 IEEE Automatic Speech Recognition and Understanding Workshop (ASRU)}, pages 1--7.

\bibitem[{Chen et~al.(2022)Chen, Zhang, Rosenberg, Ramabhadran, Moreno, Bapna, and Zen}]{Chen2022MAESTROMS}
Zhehuai Chen, Yu~Zhang, Andrew Rosenberg, Bhuvana Ramabhadran, Pedro~J. Moreno, Ankur Bapna, and Heiga Zen. 2022.
\newblock \href {https://api.semanticscholar.org/CorpusID:248006130} {Maestro: Matched speech text representations through modality matching}.
\newblock In \emph{Interspeech}.

\bibitem[{Communication et~al.(2023)Communication, Barrault, Chung, Meglioli, Dale, Dong, Duquenne, Elsahar, Gong, Heffernan, Hoffman, Klaiber, Li, Licht, Maillard, Rakotoarison, Sadagopan, Wenzek, Ye, Akula, Chen, Hachem, Ellis, Gonzalez, Haaheim, Hansanti, Howes, Huang, Hwang, Inaguma, Jain, Kalbassi, Kallet, Kulikov, Lam, Li, Ma, Mavlyutov, Peloquin, Ramadan, Ramakrishnan, Sun, Tran, Tran, Tufanov, Vogeti, Wood, Yang, Yu, Andrews, Balioglu, Costa-jussà, Celebi, Elbayad, Gao, Guzmán, Kao, Lee, Mourachko, Pino, Popuri, Ropers, Saleem, Schwenk, Tomasello, Wang, Wang, and Wang}]{seamlessm4tmassivelymultilingual}
Seamless Communication, Loïc Barrault, Yu-An Chung, Mariano~Cora Meglioli, David Dale, Ning Dong, Paul-Ambroise Duquenne, Hady Elsahar, Hongyu Gong, Kevin Heffernan, John Hoffman, Christopher Klaiber, Pengwei Li, Daniel Licht, Jean Maillard, Alice Rakotoarison, Kaushik~Ram Sadagopan, Guillaume Wenzek, Ethan Ye, Bapi Akula, Peng-Jen Chen, Naji~El Hachem, Brian Ellis, Gabriel~Mejia Gonzalez, Justin Haaheim, Prangthip Hansanti, Russ Howes, Bernie Huang, Min-Jae Hwang, Hirofumi Inaguma, Somya Jain, Elahe Kalbassi, Amanda Kallet, Ilia Kulikov, Janice Lam, Daniel Li, Xutai Ma, Ruslan Mavlyutov, Benjamin Peloquin, Mohamed Ramadan, Abinesh Ramakrishnan, Anna Sun, Kevin Tran, Tuan Tran, Igor Tufanov, Vish Vogeti, Carleigh Wood, Yilin Yang, Bokai Yu, Pierre Andrews, Can Balioglu, Marta~R. Costa-jussà, Onur Celebi, Maha Elbayad, Cynthia Gao, Francisco Guzmán, Justine Kao, Ann Lee, Alexandre Mourachko, Juan Pino, Sravya Popuri, Christophe Ropers, Safiyyah Saleem, Holger Schwenk, Paden Tomasello, Changhan Wang, Jeff
  Wang, and Skyler Wang. 2023.
\newblock \href {https://arxiv.org/abs/2308.11596} {Seamlessm4t: Massively multilingual \& multimodal machine translation}.
\newblock \emph{Preprint}, arXiv:2308.11596.

\bibitem[{Devlin et~al.(2019)Devlin, Chang, Lee, and Toutanova}]{devlin2019bertpretrainingdeepbidirectional}
Jacob Devlin, Ming-Wei Chang, Kenton Lee, and Kristina Toutanova. 2019.
\newblock \href {https://arxiv.org/abs/1810.04805} {Bert: Pre-training of deep bidirectional transformers for language understanding}.
\newblock \emph{Preprint}, arXiv:1810.04805.

\bibitem[{Gabler et~al.(2023)Gabler, Geiger, Schuppler, and Kern}]{gabler2023reconsidering}
Philipp Gabler, Bernhard~C Geiger, Barbara Schuppler, and Roman Kern. 2023.
\newblock Reconsidering read and spontaneous speech: Causal perspectives on the generation of training data for automatic speech recognition.
\newblock \emph{Information}, 14(2):137.

\bibitem[{Gala et~al.(2023)Gala, Chitale, Raghavan, Gumma, Doddapaneni, M, Nawale, Sujatha, Puduppully, Raghavan, Kumar, Khapra, Dabre, and Kunchukuttan}]{gala2023indictrans}
Jay Gala, Pranjal~A Chitale, A~K Raghavan, Varun Gumma, Sumanth Doddapaneni, Aswanth~Kumar M, Janki~Atul Nawale, Anupama Sujatha, Ratish Puduppully, Vivek Raghavan, Pratyush Kumar, Mitesh~M Khapra, Raj Dabre, and Anoop Kunchukuttan. 2023.
\newblock \href {https://openreview.net/forum?id=vfT4YuzAYA} {Indictrans2: Towards high-quality and accessible machine translation models for all 22 scheduled indian languages}.
\newblock \emph{Transactions on Machine Learning Research}.

\bibitem[{Hendrycks and Gimpel(2023)}]{hendrycks2023gaussianerrorlinearunits}
Dan Hendrycks and Kevin Gimpel. 2023.
\newblock \href {https://arxiv.org/abs/1606.08415} {Gaussian error linear units (gelus)}.
\newblock \emph{Preprint}, arXiv:1606.08415.

\bibitem[{Houlsby et~al.(2019)Houlsby, Giurgiu, Jastrzebski, Morrone, Laroussilhe, Gesmundo, Attariyan, and Gelly}]{PETNLP}
Neil Houlsby, Andrei Giurgiu, Stanislaw Jastrzebski, Bruna Morrone, Quentin Laroussilhe, Andrea Gesmundo, Mona Attariyan, and Sylvain Gelly. 2019.
\newblock Parameter-efficient transfer learning for nlp.

\bibitem[{Javed et~al.(2024{\natexlab{a}})Javed, Nawale, George, Joshi, Bhogale, Mehendale, Sethi, Ananthanarayanan, Faquih, Palit, Ravishankar, Sukumaran, Panchagnula, Murali, Gandhi, R, M, Vaijayanthi, Karunganni, Kumar, and Khapra}]{javed-etal-2024-indicvoices}
Tahir Javed, Janki Nawale, Eldho George, Sakshi Joshi, Kaushal Bhogale, Deovrat Mehendale, Ishvinder Sethi, Aparna Ananthanarayanan, Hafsah Faquih, Pratiti Palit, Sneha Ravishankar, Saranya Sukumaran, Tripura Panchagnula, Sunjay Murali, Kunal Gandhi, Ambujavalli R, Manickam M, C~Vaijayanthi, Krishnan Karunganni, Pratyush Kumar, and Mitesh Khapra. 2024{\natexlab{a}}.
\newblock \href {https://aclanthology.org/2024.findings-acl.639} {{I}ndic{V}oices: Towards building an inclusive multilingual speech dataset for {I}ndian languages}.
\newblock In \emph{Findings of the Association for Computational Linguistics ACL 2024}, pages 10740--10782, Bangkok, Thailand and virtual meeting. Association for Computational Linguistics.

\bibitem[{Javed et~al.(2024{\natexlab{b}})Javed, Nawale, George, Joshi, Bhogale, Mehendale, Sethi, Ananthanarayanan, Faquih, Palit, Ravishankar, Sukumaran, Panchagnula, Murali, Gandhi, R, M, Vaijayanthi, Karunganni, Kumar, and Khapra}]{javed2024indicvoicesbuildinginclusivemultilingual}
Tahir Javed, Janki~Atul Nawale, Eldho~Ittan George, Sakshi Joshi, Kaushal~Santosh Bhogale, Deovrat Mehendale, Ishvinder~Virender Sethi, Aparna Ananthanarayanan, Hafsah Faquih, Pratiti Palit, Sneha Ravishankar, Saranya Sukumaran, Tripura Panchagnula, Sunjay Murali, Kunal~Sharad Gandhi, Ambujavalli R, Manickam~K M, C~Venkata Vaijayanthi, Krishnan Srinivasa~Raghavan Karunganni, Pratyush Kumar, and Mitesh~M Khapra. 2024{\natexlab{b}}.
\newblock \href {https://arxiv.org/abs/2403.01926} {Indicvoices: Towards building an inclusive multilingual speech dataset for indian languages}.
\newblock \emph{Preprint}, arXiv:2403.01926.

\bibitem[{Kakwani et~al.(2020)Kakwani, Kunchukuttan, Golla, N.C., Bhattacharyya, Khapra, and Kumar}]{kakwani2020indicnlpsuite}
Divyanshu Kakwani, Anoop Kunchukuttan, Satish Golla, Gokul N.C., Avik Bhattacharyya, Mitesh~M. Khapra, and Pratyush Kumar. 2020.
\newblock {IndicNLPSuite: Monolingual Corpora, Evaluation Benchmarks and Pre-trained Multilingual Language Models for Indian Languages}.
\newblock In \emph{Findings of EMNLP}.

\bibitem[{Kessler et~al.(2021)Kessler, Thomas, and Karout}]{Kessler2021Continualwav2vec2AA}
Samuel Kessler, Bethan Thomas, and Salah Karout. 2021.
\newblock \href {https://api.semanticscholar.org/CorpusID:236469084} {Continual-wav2vec2: an application of continual learning for self-supervised automatic speech recognition}.
\newblock \emph{ArXiv}, abs/2107.13530.

\bibitem[{Mittal et~al.(2023)Mittal, Sarawagi, and Jyothi}]{mittal2023insitu}
Ashish Mittal, Sunita Sarawagi, and Preethi Jyothi. 2023.
\newblock \href {https://openreview.net/forum?id=T2Ncx_PN2K} {In-situ text-only adaptation of speech models with low-overhead speech imputations}.
\newblock In \emph{The Eleventh International Conference on Learning Representations}.

\bibitem[{Patil et~al.(2022)Patil, Singh, and Joshi}]{Patil2022UnderstandingMF}
Omkar Patil, Rahul Singh, and Tarun Joshi. 2022.
\newblock \href {https://api.semanticscholar.org/CorpusID:249097807} {Understanding metrics for paraphrasing}.
\newblock \emph{ArXiv}, abs/2205.13119.

\bibitem[{Radford et~al.(2022)Radford, Kim, Xu, Brockman, McLeavey, and Sutskever}]{radford2022whisper}
Alec Radford, Jong~Wook Kim, Tao Xu, Greg Brockman, Christine McLeavey, and Ilya Sutskever. 2022.
\newblock \href {https://doi.org/10.48550/ARXIV.2212.04356} {Robust speech recognition via large-scale weak supervision}.
\newblock \emph{arXiv preprint}.

\bibitem[{Rubenstein et~al.(2023)Rubenstein, Asawaroengchai, Nguyen, Bapna, Borsos, de~Chaumont~Quitry, Chen, Badawy, Han, Kharitonov, Muckenhirn, Padfield, Qin, Rozenberg, Sainath, Schalkwyk, Sharifi, Ramanovich, Tagliasacchi, Tudor, Velimirović, Vincent, Yu, Wang, Zayats, Zeghidour, Zhang, Zhang, Zilka, and Frank}]{rubenstein2023audiopalmlargelanguagemodel}
Paul~K. Rubenstein, Chulayuth Asawaroengchai, Duc~Dung Nguyen, Ankur Bapna, Zalán Borsos, Félix de~Chaumont~Quitry, Peter Chen, Dalia~El Badawy, Wei Han, Eugene Kharitonov, Hannah Muckenhirn, Dirk Padfield, James Qin, Danny Rozenberg, Tara Sainath, Johan Schalkwyk, Matt Sharifi, Michelle~Tadmor Ramanovich, Marco Tagliasacchi, Alexandru Tudor, Mihajlo Velimirović, Damien Vincent, Jiahui Yu, Yongqiang Wang, Vicky Zayats, Neil Zeghidour, Yu~Zhang, Zhishuai Zhang, Lukas Zilka, and Christian Frank. 2023.
\newblock \href {https://arxiv.org/abs/2306.12925} {Audiopalm: A large language model that can speak and listen}.
\newblock \emph{Preprint}, arXiv:2306.12925.

\bibitem[{Shen et~al.(2022{\natexlab{a}})Shen, Jiang, Liu, and Shi}]{Shen2022RevisitingTE}
Lingfeng Shen, Haiyun Jiang, Lemao Liu, and Shuming Shi. 2022{\natexlab{a}}.
\newblock \href {https://api.semanticscholar.org/CorpusID:246904511} {Revisiting the evaluation metrics of paraphrase generation}.
\newblock \emph{ArXiv}, abs/2202.08479.

\bibitem[{Shen et~al.(2022{\natexlab{b}})Shen, Liu, Jiang, and Shi}]{shen-etal-2022-evaluation}
Lingfeng Shen, Lemao Liu, Haiyun Jiang, and Shuming Shi. 2022{\natexlab{b}}.
\newblock \href {https://doi.org/10.18653/v1/2022.emnlp-main.208} {On the evaluation metrics for paraphrase generation}.
\newblock In \emph{Proceedings of the 2022 Conference on Empirical Methods in Natural Language Processing}, pages 3178--3190, Abu Dhabi, United Arab Emirates. Association for Computational Linguistics.

\bibitem[{Sikasote et~al.(2023)Sikasote, Siaminwe, Mwape, Zulu, Phiri, Phiri, Zulu, Nyirenda, and Anastasopoulos}]{sikasote23_interspeech}
Claytone Sikasote, Kalinda Siaminwe, Stanly Mwape, Bangiwe Zulu, Mofya Phiri, Martin Phiri, David Zulu, Mayumbo Nyirenda, and Antonios Anastasopoulos. 2023.
\newblock \href {https://doi.org/10.21437/Interspeech.2023-1979} {{Zambezi Voice: A Multilingual Speech Corpus for Zambian Languages}}.
\newblock In \emph{Proc. INTERSPEECH 2023}, pages 3984--3988.

\bibitem[{Team et~al.(2022)Team, Costa-jussà, Cross, Çelebi, Elbayad, Heafield, Heffernan, Kalbassi, Lam, Licht, Maillard, Sun, Wang, Wenzek, Youngblood, Akula, Barrault, Gonzalez, Hansanti, Hoffman, Jarrett, Sadagopan, Rowe, Spruit, Tran, Andrews, Ayan, Bhosale, Edunov, Fan, Gao, Goswami, Guzmán, Koehn, Mourachko, Ropers, Saleem, Schwenk, and Wang}]{nllbteam2022languageleftbehindscaling}
NLLB Team, Marta~R. Costa-jussà, James Cross, Onur Çelebi, Maha Elbayad, Kenneth Heafield, Kevin Heffernan, Elahe Kalbassi, Janice Lam, Daniel Licht, Jean Maillard, Anna Sun, Skyler Wang, Guillaume Wenzek, Al~Youngblood, Bapi Akula, Loic Barrault, Gabriel~Mejia Gonzalez, Prangthip Hansanti, John Hoffman, Semarley Jarrett, Kaushik~Ram Sadagopan, Dirk Rowe, Shannon Spruit, Chau Tran, Pierre Andrews, Necip~Fazil Ayan, Shruti Bhosale, Sergey Edunov, Angela Fan, Cynthia Gao, Vedanuj Goswami, Francisco Guzmán, Philipp Koehn, Alexandre Mourachko, Christophe Ropers, Safiyyah Saleem, Holger Schwenk, and Jeff Wang. 2022.
\newblock \href {https://arxiv.org/abs/2207.04672} {No language left behind: Scaling human-centered machine translation}.
\newblock \emph{Preprint}, arXiv:2207.04672.

\bibitem[{Tomanek et~al.(2024)Tomanek, Tobin, Venugopalan, Cave, Seaver, Green, and Heywood}]{10447177}
Katrin Tomanek, Jimmy Tobin, Subhashini Venugopalan, Richard Cave, Katie Seaver, Jordan~R. Green, and Rus Heywood. 2024.
\newblock \href {https://doi.org/10.1109/ICASSP48485.2024.10447177} {Large language models as a proxy for human evaluation in assessing the comprehensibility of disordered speech transcription}.
\newblock In \emph{ICASSP 2024 - 2024 IEEE International Conference on Acoustics, Speech and Signal Processing (ICASSP)}, pages 10846--10850.

\bibitem[{Vaswani(2017)}]{vaswani2017attention}
A~Vaswani. 2017.
\newblock Attention is all you need.
\newblock \emph{Advances in Neural Information Processing Systems}.

\bibitem[{Vuong et~al.(2023)Vuong, Mundnich, Bekal, Elluru, Ronanki, and Bodapati}]{vuong-etal-2023-adabert}
Tyler Vuong, Karel Mundnich, Dhanush Bekal, Veera Elluru, Srikanth Ronanki, and Sravan Bodapati. 2023.
\newblock \href {https://doi.org/10.18653/v1/2023.emnlp-industry.35} {{A}da{BERT}-{CTC}: Leveraging {BERT}-{CTC} for text-only domain adaptation in {ASR}}.
\newblock In \emph{Proceedings of the 2023 Conference on Empirical Methods in Natural Language Processing: Industry Track}, pages 364--371, Singapore. Association for Computational Linguistics.

\bibitem[{Yu et~al.(2023)Yu, Huang, Qi, and Zhou}]{Yu2023TrainingWT}
Yijiong Yu, Yongfeng Huang, Zhixiao Qi, and Zhe Zhou. 2023.
\newblock \href {https://api.semanticscholar.org/CorpusID:266359773} {Training with"paraphrasing the original text"improves long-context performance}.

\bibitem[{Zhang et~al.(2023)Zhang, Li, Zhang, Zhan, Wang, Zhou, and Qiu}]{zhang2023speechgptempoweringlargelanguage}
Dong Zhang, Shimin Li, Xin Zhang, Jun Zhan, Pengyu Wang, Yaqian Zhou, and Xipeng Qiu. 2023.
\newblock \href {https://arxiv.org/abs/2305.11000} {Speechgpt: Empowering large language models with intrinsic cross-modal conversational abilities}.
\newblock \emph{Preprint}, arXiv:2305.11000.

\bibitem[{Zhang et~al.(2020)Zhang, Kishore, Wu, Weinberger, and Artzi}]{zhang2020bertscoreevaluatingtextgeneration}
Tianyi Zhang, Varsha Kishore, Felix Wu, Kilian~Q. Weinberger, and Yoav Artzi. 2020.
\newblock \href {https://arxiv.org/abs/1904.09675} {Bertscore: Evaluating text generation with bert}.
\newblock \emph{Preprint}, arXiv:1904.09675.

\bibitem[{Üstün et~al.(2024)Üstün, Aryabumi, Yong, Ko, D'souza, Onilude, Bhandari, Singh, Ooi, Kayid, Vargus, Blunsom, Longpre, Muennighoff, Fadaee, Kreutzer, and Hooker}]{ayamodelinstructionfinetuned}
Ahmet Üstün, Viraat Aryabumi, Zheng-Xin Yong, Wei-Yin Ko, Daniel D'souza, Gbemileke Onilude, Neel Bhandari, Shivalika Singh, Hui-Lee Ooi, Amr Kayid, Freddie Vargus, Phil Blunsom, Shayne Longpre, Niklas Muennighoff, Marzieh Fadaee, Julia Kreutzer, and Sara Hooker. 2024.
\newblock \href {https://arxiv.org/abs/2402.07827} {Aya model: An instruction finetuned open-access multilingual language model}.
\newblock \emph{Preprint}, arXiv:2402.07827.

\end{thebibliography}
\appendix
\section*{Appendix}
\section{Thresholds for \ampstau}
\label{sec:thresh}
Table \ref{tabthresh} contains the iteratively obtained best thresholds for the training sets for our experiments. In case of inconsistency between different metrics, the best threshold was chosen using the validation WER for the pure ASR system.   
\begin{table}[h]
\centering
\begin{adjustbox}{max width=0.5\textwidth}
\begin{small}
\begin{tabular}{l|c|c|c|c|c}
\hline
\multirow{2}{*}{Language} &
  
\multicolumn{1}{|c|}{\multirow{1}{*}{Read}} &
\multicolumn{1}{c|}{\multirow{1}{*}{Mixed}} &
  \multicolumn{1}{c|}{\multirow{1}{*}{Mixed}} &
  \multicolumn{1}{c|}{\multirow{1}{*}{Mixed}} &
  \multicolumn{1}{c}{\multirow{1}{*}{Mixed}}  
\\
 
  \multicolumn{1}{c}{}&
  \multicolumn{1}{|c|}{BT} &
   \multicolumn{1}{|c|}{BT} &
   \multicolumn{1}{|c|}{BT} &
  \multicolumn{1}{|c|}{LLM} & 
   \multicolumn{1}{|c}{Top-K BT}\\ \hline

 \multirow{2}{*}{Hours} &
  \multirow{2}{*}{<5} &
  \multirow{2}{*}{50} &
  \multirow{2}{*}{5} &
   \multirow{2}{*}{5} &
  \multirow{2}{*}{5} 
\\ 
 
   &
   &
   &
   &

   \\ \hline

\multirow{2}{*}{Marathi} &
  \multirow{2}{*}{3.5} &
  \multirow{2}{*}{3.8} &
  \multirow{2}{*}{3.6} &
  \multirow{2}{*}{-} &
  \multirow{2}{*}{3.6} 
\\ 
 
   &
   &
   &
   &

   \\ \hline
\multirow{2}{*}{Hindi} &
  \multirow{2}{*}{3.2} &
  \multirow{2}{*}{3.2} &
  \multirow{2}{*}{3.6} &
  \multirow{2}{*}{3.6} &
  \multirow{2}{*}{3.6} 
\\ 
 
   &
   &
   &
   &

   \\ \hline
   \multirow{2}{*}{Malayalam} &
  \multirow{2}{*}{3.8} &
  \multirow{2}{*}{3.8} &
  \multirow{2}{*}{3.4} &
  \multirow{2}{*}{-} &
  \multirow{2}{*}{3.4} 
\\ 
 
   &
   &
   &
   &

   \\ \hline
   \multirow{2}{*}{Kannada} &
  \multirow{2}{*}{3.8} &
  \multirow{2}{*}{3.6} &
  \multirow{2}{*}{3.4} &
  \multirow{2}{*}{-} &
  \multirow{2}{*}{3.2} 
\\ 
 
   &
   &
   &
   &

   \\ \hline
    \multirow{2}{*}{Nyanja} &
  \multirow{2}{*}{-} &
  \multirow{2}{*}{-} &
  \multirow{2}{*}{3.8} &
  \multirow{2}{*}{-} &
  \multirow{2}{*}{-} 
\\ 
 
   &
   &
   &
   &

   \\ \hline

\end{tabular}%

\end{small}
    \end{adjustbox}
    \vspace{3pt}
\caption{Iteratively obtained threshold values for all the experimental datasets for \ampstau.}
\label{tabthresh}
\end{table}

\section{\amps for SeamlessM4T}
For all our experiments, we used the SeamlessM4T medium model along with IndicVoices \cite{javed-etal-2024-indicvoices}, and Zambezi-voice \cite{sikasote23_interspeech} datasets. Both the data and the models are free and open-sourced.
\subsection{Adapting SeamlessM4T}
\label{sec:modeldets}
The SeamlessM4T (Medium) consists of 1.2B parameters. Full fine-tuning of these components using limited amounts of labeled data for low-resource languages may result in overfitting and degradation of ASR performance. To address these issues, parameter-efficient fine-tuning methods, such as the adapter framework, have become widely adopted in natural language processing tasks. Adapters have proven effective in low-resource ASR tasks, including accent and cross-lingual adaptation.

Formally, the operations performed in the $i^{\text{th}}$ speech encoder layer can be described as follows:
\vspace{-0.1cm}
\begin{align*}
    \mathbf{H} &= \mathrm{MHA}(\mathbf{h}^{i-1}, \mathbf{h}^{i-1}, \mathbf{h}^{i-1}) \nonumber \\[-2pt]
    \mathbf{C} &= \mathrm{Convolution}(\mathbf{H}) \nonumber \\[-2pt]
    \mathbf{\hat{h}^i} &= \mathrm{FFN}(\mathbf{C}) \nonumber \\[-2pt]
    \mathbf{h}^i &= \mathrm{Adapter}(\mathbf{\hat{h}}^i)
\end{align*}

Similarly, the operations in the $i^{\text{th}}$ decoder layer can be summarized as:
\vspace{-0.1cm}
\begin{align*}
    \mathbf{D} &= \mathrm{MHA}(\mathbf{d}^{i-1}, \mathbf{d}^{i-1}, \mathbf{d}^{i-1}) \nonumber \\[-2pt]
    \mathbf{\hat{D}} &= \mathrm{MHA}(\mathbf{d}^{i-1}, \mathbf{h}^{\ell}, \mathbf{h}^{\ell}) \nonumber \\[-2pt]
    \mathbf{\hat{d}^i} &= \mathrm{FFN}(\mathbf{\hat{D}}) \nonumber \\[-2pt]
    \mathbf{d}^i &= \mathrm{Adapter}(\mathbf{\hat{d}}^i)
\end{align*}

\noindent Here, $\ell$ refers to the final encoder layer, and MHA(Q, K, V) denotes the standard multi-head attention mechanism~\cite{vaswani2017attention}, where Q, K, and V are the queries, keys, and values, respectively.
\subsection{Implementation Details}
\label{sec:impl}

The architecture of the SeamlessM4T medium incorporates a speech encoder that has 12 conformer layers, while both the text encoder and text decoder consist of 12 Transformer blocks, with a model dimension of \( D_1 = 1024 \). In our experiments, adapters were introduced after each encoder conformer layer and the decoder Transformer layer. These adapters project the original \(D_1\)-dimensional features into a reduced intermediate space of dimension \(D_2\), apply a GeLU non-linear activation function~\cite{hendrycks2023gaussianerrorlinearunits}, and then project the features back to \(D_1\). The projected layer dimension on the adapters is \( D_2 = 2048 \).  The value of \(D_2\) controls the number of trainable parameters, with smaller values of \(D_2\) reducing parameter count. With \(D_2\) set to half of \(D_1\), this setup introduced 100M trainable parameters while keeping the rest of the model frozen. 

All the fine-tuning experiments were conducted using the SeamlessM4T codebase~\cite{seamlessm4tmassivelymultilingual} released by Meta AI using NVIDIA RTX A6000 GPUs. The experiments were conducted over 20 epochs, utilizing a batch size of 8 and a learning rate of \( 5 \times 10^{-6} \). All the reported results throughout this study are based on a single fixed random seed.

The paraphrase generation using IndicTrans2 and NLLB employs a beam width of 5, while Top-K and Nucleus sampling utilize \( K = 50 \) and \( P = 0.95 \), respectively.
\begin{table*}[t]
\centering
\begin{adjustbox}{max width=\textwidth}
\begin{small}
 
\begin{tabular}{l|cccc|c}
\hline
\multirow{2}{*}{Language} &
  
\multicolumn{1}{|c}{\multirow{2}{*}{ASR}} &
\multicolumn{1}{c}{\multirow{2}{*}{\ampstau}} &
  \multicolumn{1}{c|}{\multirow{2}{*}{Meaning}} &
  \multicolumn{1}{c}{\multirow{2}{*}{Explanation}}  
\\
 
  \multicolumn{1}{c}{}&
  \multicolumn{1}{|c}{} &
   \multicolumn{1}{c}{} &
   \multicolumn{1}{c|}{} &

   \multicolumn{1}{|c}{}\\ \hline

\multirow{6}{*}{Marathi} &
  \multirow{2}{*}{aaiskrim} &
  
  \multirow{2}{*}{aayskrim} & 
   \multirow{2}{*}{icecream} &
  \multicolumn{1}{|c}{\multirow{2}{*}{Different native spelling of english word} }
\\ 
 
   &
   &
 &  &  \multicolumn{1}{|c}{ }

   \\ \cline{2-5}
   &
  \multirow{2}{*}{aplya sarkhya} &
  
  \multirow{2}{*}{aplyasarkhya} & 
   \multirow{2}{*}{like ours} &
  \multicolumn{1}{|c}{\multirow{2}{*}{Compound words joined together} }
\\ 
 
   &
   &
 &  &  \multicolumn{1}{|c}{ }

   \\ \cline{2-5}

 &
  \multirow{2}{*}{tyoob} &
  
  \multirow{2}{*}{tyub} & 
   \multirow{2}{*}{tube} &
  \multicolumn{1}{|c}{\multirow{2}{*}{Different native spelling of english word} }
\\ 
 
   &
   &
 &  &  \multicolumn{1}{|c}{ }

   \\ \hline
   
\multirow{6}{*}{Hindi} &
  \multirow{2}{*}{baaki kuch nahi} &
  
  \multirow{2}{*}{aur kuch nahi} & 
   \multirow{2}{*}{nothing else} &
  \multicolumn{1}{|c}{\multirow{2}{*}{Semantically similar phrases} }
\\ 
 
   &
   &
 &  &  \multicolumn{1}{|c}{}

   \\ \cline{2-5}
   &
  \multirow{2}{*}{bhajansangraha} &
  
  \multirow{2}{*}{bhajan sangraha} & 
   \multirow{2}{*}{prayer collection} &
  \multicolumn{1}{|c}{\multirow{2}{*}{Compound words separated} }
\\ 
 
   &
   &
 &  &  \multicolumn{1}{|c}{}

   \\ \cline{2-5}

 &
  \multirow{2}{*}{manobhavon} &
  
  \multirow{2}{*}{bhavanaon} & 
   \multirow{2}{*}{sentiments} &
  \multicolumn{1}{|c}{\multirow{2}{*}{Semantically similar words} }
\\ 
 
   &
   &
 &  &  \multicolumn{1}{|c}{}

   \\ \hline

\end{tabular}%
 
\end{small}
    \end{adjustbox}
\caption{Examples of semantically similar and linguistically different phrases and words}
\label{tabeg}
\end{table*}

\section{LLM Prompts for Paraphrasing}
\label{sec:prompt}
The paraphrasing prompt given to the Aya model for our very specific paraphrasing task has been stated below:\\
\textit{Paraphrase the following sentence in \textbf{lang}, strictly adhering to these guidelines:}
 \begin{enumerate} [itemsep=0cm]
    \item \textit{Maintain the original sentence structure and word order as much as possible.}
     \item \textit{Replace at least one word, and aim to replace as many words as feasible with Hindi synonyms or words with similar meanings.}
     \item \textit{Do not add extra words or elaborate on the description.}
     \item \textit{Preserve named entities (e.g., proper names, places) in their original form.}
     \item \textit{Convert ALL numbers to their Hindi word equivalents. This includes dates, years, percentages, and any other numerical values.}
     \item \textit{Ensure that all replacements are common Hindi words, avoiding obscure or highly technical terms.}
     \item \textit{If a direct Hindi synonym is not available, use a phrase that conveys the same meaning.}
     \item \textit{Maintain the original tense and grammatical structure of the sentence.}
     \item \textit{If the original sentence contains English words commonly used in Hindi, you may keep them unchanged.}
  
 \end{enumerate}
 
    \textit{    IMPORTANT: Double-check that NO numerical digits remain in your paraphrase. All numbers must be written out in Hindi words.}
 
     \textit{   Examples: Some Hindi examples with the required paraphrases were provided} 
     \section{Some Qualitative examples}
\subsection{Model Outputs}
\label{sec:human2}
Table \ref{tabeg} depicts examples of phrases that were acceptable for human annotation but would have incurred penalties on the use of other metrics. It can be observed that the model outputs differ from the ground truth due to native spellings of English words, whether compound words are connected or not, and semantically similar but linguistically different words and phrases. Such errors get penalized harshly by metrics like WER.

\subsection{Paraphrases}
\label{sec:para}
Table \ref{tabegnew} shows examples of sentences and their corresponding paraphrases generated via round-trip translation, where word order has been preserved to ensure semantic alignment. These were used as a guideline to create the paraphrasing prompt of the LLM. We require paraphrases where word order does not change much and where synonyms and semantically similar but linguistically different phrases are used frequently.

\begin{table*}[t]
\centering
\begin{adjustbox}{max width=\textwidth}
\begin{small}
\begin{tabular}{l|c|c}
\hline
\multirow{2}{*}{Language} &
  
\multicolumn{1}{|c|}{\multirow{2}{*}{Ground Truth}} &
  \multicolumn{1}{c}{\multirow{2}{*}{Paraphrase}}  
\\
 
  \multicolumn{1}{c}{}&
 
   \multicolumn{1}{|c|}{} &

   \multicolumn{1}{|c}{}\\ \hline

\multirow{4}{*}{Marathi} &

   \multirow{2}{*}{plij mala sagla informashun dya} &
  \multicolumn{1}{|c}{\multirow{2}{*}{krupaya tumhi mala sarva mahiti dya } }
\\ 
 
   &
 
  &  \multicolumn{1}{|c}{}

   \\ \cline{2-3}

 &
  \multirow{2}{*}{aani ashya bimarina rokhne} &

  \multicolumn{1}{|c}{\multirow{2}{*}{aani ashya roganpasun bachav karne} }
\\ 
 
   &
   &
   \multicolumn{1}{|c}{}

   \\ \hline
   
\multirow{4}{*}{Hindi} &
  \multirow{2}{*}{draiving karte samay mobail fon ka yuj nahi kare} &

  \multicolumn{1}{|c}{\multirow{2}{*}{gaadi chalate samay mobail fon ka upyog na kare } }
\\ 
 
   &
   \multicolumn{1}{|c}{} &
  \multicolumn{1}{|c}{}

   \\ \cline{2-3}

 &
  \multirow{2}{*}{kareer banana pasand karunga iska pramukh kaaran} &

  \multicolumn{1}{|c}{\multirow{2}{*}{kareer banana chahunga jiska mukhya kaaran} }
\\ 
 
   &
  \multicolumn{1}{|c}{} &
  \multicolumn{1}{|c}{}

   \\ \hline

\end{tabular}%

\end{small}
    \end{adjustbox}
    \vspace{3pt}
\caption{Examples demonstrating the ideal paraphrases for \amps.}
\vspace{-7pt}
\label{tabegnew}
\end{table*}
\section{Paraphrase Evaluation Metrics}
\label{sec:metrics}

\begin{enumerate}
    \item \textbf{Word Error Rate (WER)} measures the number of mistakes in transcription as a ratio of the number of words. These errors could be substitutions, insertions or deletions.
    \begin{equation} 
    \text{WER}=\frac{\text{Substitutions+Inclusions+Deletions}}{\text{Words in Reference Text}}
    \end{equation}
    \item \textbf{METEOR} \cite{banerjee-lavie-2005-meteor} is used for evaluating of machine translation quality. It has also previously been used for evaluating paraphrase quality\cite{shen-etal-2022-evaluation}. It aligns words in the candidate and reference translations based on word level matches, including same meaning words and stemming.
    
    \item \textbf{BERTScore} \cite{zhang2020bertscoreevaluatingtextgeneration} evaluates the similarity between two texts by using  BERT embeddings\cite{devlin2019bertpretrainingdeepbidirectional} (Bidirectional Encoder Representations from Transformers). It  captures contextual meaning and semantics by computing the cosine similarity between token embeddings from a reference sentence and a candidate sentence. We used AI4Bharat's IndicBERT \cite{kakwani2020indicnlpsuite}for our BERTScores.
    
    \item \textbf{Other metrics} like PARAScore \cite{shen-etal-2022-evaluation}, BBScore \cite{Shen2022RevisitingTE}, LATTEScore \cite{10447177} and ROUGE \cite{Patil2022UnderstandingMF} have been used in the past for evaluation of paraphrases.

\end{enumerate}

\section{\amps for Read Speech}
\label{sec:more1}
Table \ref{tabread} depicts \amps for Marathi, Malayalam, and Kannada using all the read speech of the IndicVoices~\cite{javed-etal-2024-indicvoices} dataset. Training sets of Kannada, Malayalam, and Marathi were of duration 2.64, 2.01, and 4.84, respectively. All validation sets were of a half-hour duration. It can be observed that \ampstau performs the best for Marathi, Malayalam, and Kannada round-trip translated read speech.

\begin{table}[h]
\centering
\begin{adjustbox}{max width=0.48\textwidth}
\begin{small}
 
\begin{tabular}{l|c|c|ccc}
\hline
\multirow{4}{*}{Language} &
\multicolumn{1}{|c|}{\multirow{1}{*}{Paraphrase}} &
  \multicolumn{1}{|c|}{\multirow{2}{*}{Baseline}} &
  \multicolumn{3}{c}{\multirow{1}{*}{Read Speech}} 
\\
 
  \multicolumn{1}{c}{}&
  \multicolumn{1}{|c|}{Type} &
  \multicolumn{1}{|c|}{} &
  \multicolumn{3}{c}{RT Trans} \\ 

\cline{2-6}
 &
  \multicolumn{1}{|l|}{\multirow{2}{*}{Configuration}} &
  \multicolumn{1}{c|}{\multirow{2}{*}{-}} &
    \multicolumn{1}{c}{\multirow{2}{*}{ASR}} &
    \multicolumn{1}{c}{\multirow{2}{*}{\amps}} &
    \multicolumn{1}{c}{\multirow{2}{*}{\ampstau}}

\\ &
  \multicolumn{1}{|l|}{} &
\multicolumn{1}{|c|}{} &
\multicolumn{1}{c}{} &
\multicolumn{1}{c}{} &
\multicolumn{1}{c}{} 
 
   \\ \hline
\multirow{6}{*}{Marathi} &
  \multirow{2}{*}{WER $\downarrow$} &
  \multirow{2}{*}{38.65} &
  \multirow{2}{*}{34.04} &
  \multirow{2}{*}{32.30} &
  \multirow{2}{*}{\textbf{31.25}} 
\\
 &
   &
   &
   &
   &
   \\ \cline{2-6}
   &
  \multirow{2}{*}{METEOR $\uparrow$} &
  \multirow{2}{*}{59.84} &
  \multirow{2}{*}{67.26} &
  \multirow{2}{*}{68.83} &
  \multirow{2}{*}{\textbf{70.04}} 
\\
 &
   &
   &
   &
   &
   \\ \cline{2-6}
   &
  \multirow{2}{*}{BERTScore $\uparrow$} &
  \multirow{2}{*}{81.01} &
  \multirow{2}{*}{87.71} &
  \multirow{2}{*}{88.65} &
  \multirow{2}{*}{\textbf{89.18}} 
\\
 &
   &
   &
   &
   &
  \\ \hline

   \multirow{6}{*}{Malayalam} &
  \multirow{2}{*}{WER $\downarrow$} &
  \multirow{2}{*}{56.15} &
  \multirow{2}{*}{55.38} &
  \multirow{2}{*}{55.17} &
  \multirow{2}{*}{\textbf{54.58}} 
\\
 &
   &
   &
   &
   &
  \\ \cline{2-6}
   &
  \multirow{2}{*}{METEOR $\uparrow$} &
  \multirow{2}{*}{43.69} &
  \multirow{2}{*}{45.85} &
  \multirow{2}{*}{45.59} &
  \multirow{2}{*}{\textbf{46.22}}
\\
 &
   &
   &
   &
   &
   \\ \cline{2-6}
   &
  \multirow{2}{*}{BERTScore $\uparrow$} &
  \multirow{2}{*}{84.35} &
  \multirow{2}{*}{85.72} &
  \multirow{2}{*}{\textbf{86.01}} &
  \multirow{2}{*}{85.99} 
\\
 &
   &
   &
   &
   &
   \\ \hline
   \multirow{6}{*}{Kannada} &
  \multirow{2}{*}{WER $\downarrow$} &
  \multirow{2}{*}{69.29} &
  \multirow{2}{*}{61.86} &
  \multirow{2}{*}{61.3} &
  \multirow{2}{*}{\textbf{59.64}} 
\\
 &
   &
   &
   &
   &
   \\ \cline{2-6}
   &
  \multirow{2}{*}{METEOR $\uparrow$} &
  \multirow{2}{*}{31.13} &
  \multirow{2}{*}{38.95} &
  \multirow{2}{*}{39.80} &
  \multirow{2}{*}{\textbf{40.63}}
\\
 &
   &
   &
   &
   &
  \\ \cline{2-6}
   &
  \multirow{2}{*}{BERTScore $\uparrow$} &
  \multirow{2}{*}{76.65} &
  \multirow{2}{*}{82.48} &
  \multirow{2}{*}{82.52} &
  \multirow{2}{*}{\textbf{83.04}} 
\\
 &
   &
   &
   &
   &
    \\ \hline

\end{tabular}%
 
\end{small}
    \end{adjustbox}
    \vspace{3pt}
\caption{Comparison of ASR performance for pure ASR, \amps and \ampstau with round-trip translated (RT Trans) read-speech data for Marathi, Malayalam and Kannada}

\label{tabread}
\end{table}

\section{5 hour \amps for Other languages}
\label{sec:more2}
Table \ref{tabparatype} depicts the two different round-trip translation methods used for \amps for 5 hours each of mixed Marathi, Malayalam and Kannada speech. It can be observed that the two methods have comparable performance, with normal round-trip translation performing slightly better than the top-K and nucleus (top-P) setting.

\begin{table}[h]
\centering
\begin{adjustbox}{max width=0.48\textwidth}
\begin{small}
\renewcommand{\arraystretch}{1.2}

\begin{tabular}{l|c|c|cc|cc}
\hline
\multirow{4}{*}{Language} &
\multicolumn{1}{|c|}{\multirow{1}{*}{Paraphrase}} &

\multicolumn{1}{|c|}{\multirow{2}{*}{-}} &
  \multicolumn{2}{c|}{\multirow{1}{*}{Mixed Speech}} &
 
  \multicolumn{2}{c}{\multirow{1}{*}{Mixed Speech}}
\\
 
  \multicolumn{1}{c}{}&
  \multicolumn{1}{|c|}{Type} &
 \multicolumn{1}{|c|}{} &

  \multicolumn{2}{c|}{RT Trans} &

   \multicolumn{2}{c}{TK+Nuc RT Trans}\\ 

\cline{2-7}
 &
  \multicolumn{1}{|l|}{\multirow{2}{*}{Configuration}} &

 \multicolumn{1}{c|}{\multirow{2}{*}{ASR}} &
    \multicolumn{1}{c}{\multirow{2}{*}{\amps}} &
    \multicolumn{1}{c|}{\multirow{2}{*}{\ampstau}} &

    \multicolumn{1}{c}{\multirow{2}{*}{\amps}} &
    \multicolumn{1}{c}{\multirow{2}{*}{\ampstau}} 
 
\\ &
  \multicolumn{1}{|l|}{} &

\multicolumn{1}{c|}{} &
\multicolumn{1}{c}{} &
\multicolumn{1}{c|}{} &

\multicolumn{1}{c}{} &
\multicolumn{1}{c}{} 
 
   \\ \hline
\multirow{6}{*}{Marathi} &
  \multirow{2}{*}{WER $\downarrow$} &

  \multirow{2}{*}{24.70} &
  \multirow{2}{*}{\textbf{24.44}} &
  \multirow{2}{*}{24.60} &

  \multirow{2}{*}{\textbf{24.56}} &
  \multirow{2}{*}{24.75}
\\
 &
   &
   &
   &
   &
   &

   \\ \cline{2-7}
 

   &
  \multirow{2}{*}{METEOR $\uparrow$} &
 
  \multirow{2}{*}{76.66} &
  \multirow{2}{*}{76.80} &
  \multirow{2}{*}{\textbf{77.11}} &

  \multirow{2}{*}{76.50} &
  \multirow{2}{*}{\textbf{76.74}}
\\
 &
   &
   &
   &
   &
   &
   \\ \cline{2-7}
   &
  \multirow{2}{*}{BERTScore $\uparrow$} &

  \multirow{2}{*}{91.77} &
  \multirow{2}{*}{91.83} &
  \multirow{2}{*}{\textbf{92.01}} &

  \multirow{2}{*}{91.59} &
  \multirow{2}{*}{\textbf{91.83}}
\\
 &
   &
   &
   &
   &
   &
 \\ \hline
   \multirow{6}{*}{Malayalam} &
  \multirow{2}{*}{WER $\downarrow$} &

  \multirow{2}{*}{47.90} &
  \multirow{2}{*}{47.11} &
  \multirow{2}{*}{\textbf{46.06}} &

  \multirow{2}{*}{46.41} &
  \multirow{2}{*}{\textbf{46.27}}
\\
 &
   &
   &
   &
   &
   &
   \\ \cline{2-7}
 

   &
  \multirow{2}{*}{METEOR $\uparrow$} &
 
  \multirow{2}{*}{55.29} &
  \multirow{2}{*}{\textbf{55.86}} &
 
  \multirow{2}{*}{55.82} &
  \multirow{2}{*}{56.84} &
 
  \multirow{2}{*}{\textbf{56.92}}
\\
 &
   &
   &
   &
   &
   &
  \\ \cline{2-7}
   &
  \multirow{2}{*}{BERTScore $\uparrow$} &
 
  \multirow{2}{*}{89.82} &
  \multirow{2}{*}{\textbf{90.18}} &
  
  \multirow{2}{*}{89.96} &
  \multirow{2}{*}{\textbf{90.27}} &

  \multirow{2}{*}{90.25}
\\
 &
   &
   &
   &
   &
   &
   \\ \hline
   \multirow{6}{*}{Kannada} &
  \multirow{2}{*}{WER $\downarrow$} &
 
  \multirow{2}{*}{46.77} &
  \multirow{2}{*}{46.53} &
  \multirow{2}{*}{\textbf{46.35}} &

  \multirow{2}{*}{46.24} &
  \multirow{2}{*}{\textbf{46.22}}
\\
 &
   &
   &
   &

   &
   &
   \\ \cline{2-7}


   &
  \multirow{2}{*}{METEOR $\uparrow$} &
 
  \multirow{2}{*}{53.77} &
  \multirow{2}{*}{54.49} &
 
  \multirow{2}{*}{\textbf{54.80}} &
  \multirow{2}{*}{54.34} &
  
  \multirow{2}{*}{\textbf{54.47}}
\\
 &
   &
   &
  
   &
   &
   &
  \\ \cline{2-7}
   &
  \multirow{2}{*}{BERTScore $\uparrow$} &

  \multirow{2}{*}{87.90} &
  \multirow{2}{*}{87.78} &
  \multirow{2}{*}{\textbf{87.92}} &

  \multirow{2}{*}{87.86} &
  \multirow{2}{*}{\textbf{87.99}}
\\
 &
   &
   &
 
   &
   &
   \\ \hline

\end{tabular}%
 
\end{small}
    \end{adjustbox}
    \vspace{3pt}
\caption{Comparison of ASR performance for pure ASR, \amps and \ampstau for normal round-trip translated (RT Trans) and top K + Nucleus sampled round-trip translated (TK+Nuc RT Trans) mixed data for Marathi, Malayalam, and Kannada}

\label{tabparatype}

\end{table}
\section{Details of Human Evaluation}
\label{sec:human1}
Human evaluation was outsourced to an annotation company based in India, and INR 45 was paid for every audio. Each sentence was given a maximum score of 5 for perfect transcription. In cases of erroneous transcriptions, 0.5 points were deducted for every instance of a minor error, and 1 point was deducted for every instance of a major error. Minor errors included small character errors or tense changes that led to wrong grammar. Major errors included wrong transcriptions, missed words, and wrongly spelled native words. The annotators were instructed to give no penalty for incomprehensible audio, varying native spellings of English words or proper nouns, semantically similar but linguistically different words, and broken or connected compound words.

\section{Paraphrase Supervision for Purely Speech-to-Text Models}
\label{sec:pure}
To provide a comparison for our multimodal model technique, we propose an alternative approach involving pretraining and finetuning for purely speech-to-text ASR models. The hypothesis is that training an ASR model first on speech paired with paraphrased transcripts, followed by finetuning it on speech with original transcripts, will result in a model that is more robust to mispronunciations and noisy inputs. By learning to associate unclear or imprecise utterances with semantically similar phrases, this model should outperform one trained exclusively on ground-truth labels when evaluated on noisy test sets despite exposure to similar amounts of data. To support our hypothesis, we used the Whisper ASR model trained sequentially using paraphrased transcripts followed by the ground truth, with an ASR training objective.

\subsection{Whisper }

Whisper \cite{radford2022whisper}, developed by OpenAI, utilizes a transformer-based encoder-decoder framework suitable for a range of speech-related tasks. The model comprises an audio encoder that processes raw audio inputs, transforming them into log-mel spectrograms. This input is fed into multiple transformer layers designed to capture long-range dependencies within the audio data. The text decoder, operating autoregressively, generates transcriptions from the processed audio features while integrating task-specific tokens for seamless task-switching among any auxilliary tasks.

\subsection{Experiment and Results}
The Whisper model was trained sequentially with 5-hour round-trip translated read speech data in three different ways - training with ground truth training followed by paraphrased training, paraphrase training followed by ground truth training, and finally, ground truth training repeated twice. 

The WER (\%) values for Hindi read speech were 87.68 for direct inference, 42.33 for ground truth - ground truth training, 47.34 for paraphrase - ground truth training and 43.78 for ground truth - paraphrase training. Since pure ground truth training WER is the best, we chose not to proceed with this experiment as this strongly supports that multimodality of a model is essential for \amps.

\end{document}